\documentclass[conference]{IEEEtran}
\usepackage{cite}
\usepackage{amsmath,amssymb,amsfonts}
\usepackage{algorithmic}
\usepackage{graphicx}
\usepackage{textcomp}
\usepackage{xcolor}
\usepackage{fancyhdr}
\usepackage[utf8]{inputenc}
\usepackage{graphicx}
\usepackage{caption}
\usepackage{hyperref}
\usepackage{booktabs}
\usepackage{amsmath}
\usepackage{textcomp}
\usepackage{gensymb}
\usepackage[export]{adjustbox}
\usepackage{listings}
\usepackage{float}
\usepackage{enumitem}
\usepackage{indentfirst}
\usepackage{booktabs}
\usepackage{multirow}
\usepackage{colortbl}
\usepackage{setspace}
\usepackage{xcolor}
\usepackage{caption}
\usepackage{subcaption}
\usepackage{etoolbox}
\usepackage{graphicx}
\usepackage{tikz}
\usepackage{mathtools}
\usepackage[numbers]{natbib}
\usepackage{array}
\usepackage{hyperref}
\usepackage{pdfpages}

\definecolor{codegreen}{rgb}{0,0.6,0}
\definecolor{codegray}{rgb}{0.5,0.5,0.5}
\definecolor{codepurple}{rgb}{0.58,0,0.82}
\definecolor{backcolour}{rgb}{0.95,0.95,0.92}
\def\BibTeX{{\rm B\kern-.05em{\sc i\kern-.025em b}\kern-.08em
    T\kern-.1667em\lower.7ex\hbox{E}\kern-.125emX}}

\renewcommand{\figurename}{Gambar}
\renewcommand{\tablename}{Tabel}
\renewcommand{\refname}{Daftar Referensi}

\newcommand{\subbab}{Subbab}

\newcommand{\gambar}{Gambar}
\newcommand{\tabel}{Tabel}

\begin{document}

\title{{Pengembangan Model untuk Mendeteksi Kerusakan pada Terumbu Karang dengan Klasifikasi Citra}}

\author{
\IEEEauthorblockN{\footnotesize\textbf{Fadhil Muhammad, Alif Bintang Elfandra, Iqbal Pahlevi Amin, Alfan Wicaksono}}

\IEEEauthorblockA{\footnotesize
\texttt{\{fadhil.muhammad23, alif.bintang, iqbal.pahlevi, alfan\} @ ui.ac.id}\\
Fakultas Ilmu Komputer, Universitas Indonesia, Depok, 16424\\
\textit{Corresponding Author: Fadhil Muhammad}
}
}

\twocolumn[
\begin{@twocolumnfalse}
\maketitle
\thispagestyle{fancy}
\end{@twocolumnfalse}
\begin{abstract}
Keanekaragaman hayati terumbu karang yang melimpah di perairan Indonesia menjadi aset berharga yang perlu dilestarikan. Perubahan iklim yang cepat dan aktivitas manusia yang tidak terkendali telah mengakibatkan kerusakan ekosistem terumbu karang, termasuk pemutihan terumbu karang yang menjadi tanda kritisnya kondisi kesehatan terumbu karang. Oleh karena itu, penelitian ini bertujuan untuk mengembangkan model klasifikasi yang akurat untuk membedakan antara karang yang sehat dan karang yang mengalami pemutihan. Penelitian ini menggunakan dataset khusus yang terdiri dari $923$ gambar yang dikumpulkan dari Flickr menggunakan API Flickr. Dataset ini terdiri dari dua kelas yang berbeda, yaitu karang sehat ($438$ gambar) dan karang yang mengalami pemutihan ($485$ gambar). Gambar-gambar tersebut telah diubah ukurannya menjadi maksimum $300$ piksel untuk lebar atau tinggi, mana pun yang lebih tinggi, guna mempertahankan ukuran yang konsisten di seluruh dataset. Metode yang digunakan dalam penelitian ini adalah penggunaan model machine learning, terutama jaringan saraf konvolusi (CNN), untuk mengenali dan membedakan pola visual yang terkait dengan karang sehat dan karang yang mengalami pemutihan. Dalam hal ini, dataset ini dapat digunakan untuk melatih dan menguji model-model klasifikasi yang berbeda untuk mencapai hasil yang optimal. Dengan memanfaatkan model ResNet, didapat bahwa \textit{model from scratch} pada ResNet dapat mengungguli \textit{pretrained model} dalam segi presisi dan akurasi. Keberhasilan dalam pengembangan model klasifikasi dan yang akurat akan memberikan manfaat besar bagi peneliti dan ahli biologi laut dalam pemahaman yang lebih baik terhadap kesehatan terumbu karang. Model-model ini juga dapat digunakan untuk memantau perubahan lingkungan terumbu karang, sehingga memberikan kontribusi nyata dalam upaya konservasi dan restorasi ekosistem yang mempunyai pengaruh dalam kehidupan.
\end{abstract}

]

\section{Pendahuluan}
\label{bab:1}
Dalam pekerjaan ini, dilakukan pengembangan model berbasis \emph{data-driven} yang bertujuan untuk mendeteksi kerusakan terumbu karang dengan memanfaatkan informasi citra bawah laut. Selain sesuai dengan tema besar "\textit{Life Below Water}" yang merupakan salah satu dari 17 \textit{Sustainable Development Goals}\footnote{\url{https://sdgs.un.org/goals}}, Topik ini sangat penting dan relevan mengingat keadaan terumbu karang yang semakin mengkhawatirkan di Indonesia~\citep{mous2000cyanide}. \subbab~\ref{bab1:latar_belakang} mendiskusikan motivasi utama dari pekerjaan ini dan juga gambaran umum tentang model deteksi kerusakan terumbu karang yang diusulkan. Kemudian, \subbab~\ref{bab1:tujuan} menyampaikan tujuan dan manfaat dari pekerjaan ini. Terakhir, \subbab~\ref{bab1:scope} memberikan penjelasan tentang batasan dari masalah yang disorot pada pekerjaan ini.

\subsection{Latar Belakang}
\label{bab1:latar_belakang}
Terumbu karang merupakan ekosistem yang sangat penting dan merupakan habitat bagi sebagian besar spesies laut~\citep{gates2011nature}. \citet{fisher2015species} bahkan menyatakan bahwa habitat terumbu karang memiliki tingkat biodiversitas tertinggi dari semua ekosistem yang ada di laut. Oleh karena itu, kerusakan terumbu karang dapat mengganggu ekosistem, membunuh ikan-ikan di sekitar, dan bahkan menyebabkan kemiskinan bagi para nelayan yang sulit mencari ikan.

Indonesia memiliki kekayaan terumbu karang yang luar biasa, menjadikannya salah satu negara dengan keanekaragaman hayati terumbu karang terbesar di dunia~\citep{allen2008conservation}. 
Namun, terumbu karang Indonesia saat ini menghadapi ancaman yang serius akibat berbagai faktor, seperti perubahan iklim~\citep{hoey2016recent,munday2009climate}, polusi~\citep{loya1980effects}, penangkapan ikan yang tidak bertanggung jawab~\citep{mous2000cyanide}, dan aktivitas manusia lainnya.

Menurut Lembaga Ilmu Pengetahuan Indonesia (LIPI), hanya $6{,}56$\% terumbu karang  di Indonesia yang mempunyai kondisi sangat baik. Secara keseluruhan, terumbu karang yang dalam kategori baik mengalami penurunan tren, sementara kategori rusak mengalami peningkatan jika dibandingkan dengan tahun sebelumnya.\footnote{\url{http://ejournal-balitbang.kkp.go.id/index.php/jkpt/article/view/12066/8224}}

Salah satu masalah yang mendesak adalah pemutihan terumbu, yaitu  ketika kondisi lingkungan, seperti suhu air yang tinggi atau paparan cahaya yang berlebihan, menyebabkan terumbu karang kehilangan alga simbiotik yang mendukung pertumbuhannya~\citep{glynn1993coral}.
Pemutihan terumbu karang dapat mengakibatkan kematian massal karang dan mengurangi keragaman hayati, dengan dampak jangka panjang yang signifikan terhadap ekosistem laut dan manusia. Oleh karena itu, pemantauan secara berkala terhadap terumbu karang, khususnya terkait pemutihan, merupakan hal yang bermanfaat dan membantu menghentikan proses pemutihan ke tahap yang lebih serius. Ini adalah masalah yang diangkat pada penelitian kami.

Memantau kerusakan terumbu karang secara manual merupakan tugas yang sangat sulit dan memakan waktu. Area terumbu karang yang luas dan kompleks memerlukan upaya yang besar untuk mengidentifikasi perubahan dan kerusakan secara akurat. 
Oleh karena itu, untuk meningkatkan efisiensi dan efektivitas pemantauan, diperlukan penggunaan teknik-teknik baru seperti pemantauan otomatis menggunakan citra dari satelit~\citep{gang2006satellite} atau kapal selam tanpa awak. 

Citra-citra terumbu karang yang diperoleh melalui sistem pemantau kemudian diolah menggunakan model komputasi yang dapat memberikan gambaran secara otomatis apakah terumbu karang sedang dalam proses mengalami kerusakan atau tidak.

Dalam konteks ini, pendekatan \textit{binary classification} dapat diterapkan pada data citra yang terkumpul untuk membedakan antara area terumbu karang yang sehat dan area yang mengalami kerusakan. Dalam rangka mencapai hal tersebut, model yang memanfaatkan \textit{deep learning} dapat digunakan. Model berbasis \textit{binary classification} ini dapat dilatih menggunakan data citra terumbu karang yang dikumpulkan melalui penginderaan jauh atau pengambilan gambar bawah laut, dan model ini akan mampu mengidentifikasi pola visual yang khas dari terumbu karang yang sehat serta membedakannya dari kerusakan seperti pemutihan atau pertumbuhan alga yang berlebihan. Teknik ini memungkinkan pemantauan secara terus-menerus dan dapat memberikan hasil dengan akurasi yang lebih tinggi dibandingkan dengan metode manual.

Kami mengambil pendekatan yang berbeda dalam penelitian ini. Selain melakukan \emph{preprocessing} terhadap gambar, kami berusaha untuk mengadaptasi model yang ada sekaligus membuat model baru dari awal (\emph{model from scratch}) dan membandingkan performanya dengan model yang sudah ada. Penelitian ini juga akan mencari tahu bagaimana cara model mengklasifikasikan suatu gambar ke dalam suatu kelas.

\subsection{Tujuan dan Manfaat}
\label{bab1:tujuan}
Pemanfaatan \emph{deep learning} untuk mendeteksi terumbu karang yang rusak bertujuan untuk mengidentifikasi secara akurat terumbu karang yang mengalami kerusakan atau terancam, serta memonitor kondisi mereka secara \emph{real-time}. Dengan melatih model \emph{deep learning} menggunakan data citra atau video terumbu karang, kita dapat mengembangkan sistem yang efisien dalam mendeteksi perubahan warna, struktur yang rusak, atau tanda-tanda penyakit. Integrasi dengan kamera bawah air atau drone memungkinkan pengawasan terus-menerus dan respons cepat terhadap perubahan dalam kondisi terumbu karang, membantu dalam upaya konservasi dan pemulihan terumbu karang.

\subsection{Batasan yang Digunakan}
\label{bab1:scope}
Batasan yang digunakan antara lain \emph{dataset} yang dipakai bersumber dari Flickr. Semua gambar pada \emph{dataset} berlatar di dalam laut, dan hanya terdapat dua kelas pada \emph{dataset} yang digunakan, yaitu \emph{bleached} dan \emph{healthy}.

\section{Kajian Pustaka}
\label{bab:2}

Bagian ini akan mengulas konsep-konsep penting terkait dengan topik penelitian. \subbab~\ref{bab2:cnn} menjelaskan struktur dan komponen utama CNN dalam mempelajari representasi citra. \subbab~\ref{bab2:transferlearning} membahas penggunaan \textit{pretrained model} dari ImageNet Large Scale Visual Recognition Challenge untuk meningkatkan performa model \textit{deep learning} dalam permasalahan baru. \subbab~\ref{bab2:residualnetworks} menjelaskan arsitektur ResNet yang menggunakan \textit{shortcut connections} untuk mengatasi \textit{loss information} pada input. \subbab~\ref{bab2:gradcam} menjelaskan mengenai Grad-CAM yang digunakan untuk menjelaskan hasil prediksi model.
Bab ini juga mencakup \subbab~\ref{bab2:penelitianterdahulu} yang mengimplementasikan CNN dalam klasifikasi kondisi terumbu karang.

\subsection{\textit{Convolutional Neural Network}}
\label{bab2:cnn}
Dalam \textit{deep learning, convolutional neural network} atau dikenal dengan CNN merupakan algoritma yang digunakan untuk mempelajari representasi citra dan dapat diterapkan pada berbagai permasalahan \textit{computer vision}~\cite{MAHMOOD2017383, cnn2, cnn3}. Arsitektur CNN menggunakan operasi matematika yang disebut \textit{convolution} sebagai pengganti perkalian matrix pada setiap \textit{layer} dan dirancang untuk memproses data \textit{pixel} yang digunakan pada \textit{image processing} dan \textit{image recognition.} CNN terdiri atas tiga jenis \textit{layers} utama, yaitu \textit{convolutional layer, pooling layer,} dan \textit{fully connected layer}~\citep{cnn, cnn3}. Ketiga \textit{layers} tersebut akan ditumpuk untuk membentuk arsitektur CNN. Selain ketiga \textit{layers} tersebut, CNN juga menggunakan fungsi aktivasi ReLU~\citep{relu, cnn3} pada \textit{convolutional layer} untuk memberikan nonlinearitas pada CNN~\citep{cnn3}. Pada CNN, \textit{input layer} menerima nilai \textit{pixel} dari citra~\citep{cnn}.

\subsection{\textit{Transfer Learning}}
\label{bab2:transferlearning}
\textit{Transfer learning} merupakan metode yang digunakan untuk meningkatkan performa model \textit{deep learning} untuk menyelesaikan sebuah permasalahan baru. Metode ini memanfaatkan pengetahuan atau kemampuan model yang sudah pernah dilatih menggunakan data pada domain permasalahan yang berhubungan~\cite{5288526} atau disebut sebagai \textit{pretrained model}. Dengan memanfaatkan metode ini, tidak diperlukan untuk membuat model mulai dari awal. Metode ini dapat menghasilkan model dalam waktu yang lebih singkat dengan akurasi tinggi. Pada umumnya, \textit{pretrained model} yang digunakan adalah model yang telah dilatih menggunakan data dari ImageNet Large Scale Visual Recognition Challenge~\citep{ilsvrc}, seperti AlexNet~\citep{alexnet}, ZFNET~\citep{zfnet}, GoogleNet~\citep{googlenet}, ResNet~\citep{resnet}, SENet~\citep{senet}, dan lain-lain.

\subsection{\textit{Residual Networks}}
\label{bab2:residualnetworks}
\textit{Residual networks} atau ResNet merupakan salah satu arsitektur CNN. ResNet merupakan arsitektur yang dikembangkan oleh peniliti microsoft di Asia, yang kemudian memenangkan ImageNet Large Scale Visual Recognition Challenge pada tahun 2015. Berdasarkan penelitian yang dilakukan oleh \citet{architecture-comparison} pada data citra validasi ImageNet-1k untuk permasalahan klasifikasi, model yang mencapai top-1 dan top-5 akurasi adalah NASNet-A-Large, namun dengan kompleksitas komputasi tertinggi. Selain itu, beberapa model lain yang memiliki kompleksitas komputasi lebih rendah (di bawah $15$ G-FLOPs), namun tetap memiliki top-1 dan top-5 akurasi yang tinggi (di atas $90$\%) di antaranya adalah SE-ResNeXt-50, SE-ResNeXt-101, Inception-ResNet-v2, ResNet-50, ResNet-101, dan ResNet-152 yang merupakan keluarga algoritma ResNet.

Seperti yang dapat dilihat pada \tabel~\ref{tab:coral-texture-accuracy}, penelitian yang dilakukan oleh \citet{GOMEZRIOS2019315} pada permasalahan klasifikasi citra terumbu karang juga menunjukkan bahwa arsitektur ResNet memiliki hasil metrik akurasi yang lebih tinggi dibandingkan Inception-v3 dan DenseNet.

\begin{table}[H]
    \centering
\begin{tabular}{c c c}
\hline
& EILAT & RSMAS \\
\hline
\textbf{Inception v3} & $93{,}63$ & $89{,}59$ \\
\textbf{ResNet-50} & \textbf{$95{,}87$} & \textbf{$95{,}34$} \\
\textbf{ResNet-152} & $95{,}25$ & $94{,}66$ \\
\textbf{DenseNet-121} & $95{,}07$ & $94{,}79$ \\
\textbf{DenseNet-161} & $94{,}98$ & $94{,}52$ \\
\hline
\end{tabular}

    \caption{Komparasi akurasi arsitektur CNN pada klasifikasi tekstur terumbu karang~\cite{GOMEZRIOS2019315}.
    \label{tab:coral-texture-accuracy}}
\end{table}

Semakin rumit suatu citra, maka semakin banyak layer CNN pula yang dibutuhkan untuk mengekstrak informasi yang terdapat pada citra tersebut. Secara teori, semakin banyak jumlah layer pada CNN, maka model \textit{neural network} akan belajar lebih baik dan semakin tinggi akurasinya~\cite{resnet}. Namun, pada kenyataannya semakin banyak jumlah layer pada CNN, maka akan terjadi \textit{loss information} pada input. Hal ini disebabkan perhitungan gradien menjadi tidak efektif dan menjadi sangat kecil (\textit{vanishing gradient}) atau mungkin saja sangat besar (\textit{exploding gradient})~\cite{resnet}. 

Permasalahan tersebut dapat diatasi dengan metode \textit{deep residual learning}~\cite{resnet}. Metode ini menambahkan sebuah \textit{shortcut connections} yang dapat melewati layer tertentu untuk menghubungkan input dengan output dari \textit{convolutional layer}~\cite{resnet}. \textit{Shortcut connection} berguna untuk mencegah \textit{loss information} pada input yang melalui untaian \textit{neural networks} yang panjang. Pada arsitektur ResNet, kombinasi dari beberapa layer dan \textit{shorcut connections} disebut sebagai satu \textit{building block} yang secara matematis dapat dituliskan sebagai: 
\begin{equation}
    y = F(x, {W}) + x \, ,
\end{equation}
dengan $x$ sebagai input, $W$ adalah parameter fungsi $F$, $F(x, {W})$ sebagai hasil \textit{residual mapping} dari suatu \textit{building block}, dan $y$ sebagai output.

\subsection{Grad-CAM (Gradient-weighted Class Activation Mapping)}
\label{bab2:gradcam}
Grad-CAM merupakan teknik yang berguna untuk menjelaskan pada bagian mana model CNN melihat. Teknik ini dapat digunakan untuk menjelaskan suatu \textit{output} dari sebuah model \textit{machine learning}. Grad-CAM menggunakan informasi gradien target yang berada pada layer \textit{convolutional} terakhir model untuk menghasilkan \textit{map} lokalisasi yang menyoroti daerah penting dari input citra untuk prediksi~\citep{gradcam, cnn2}. Selain dapat menjelaskan aktivasi pada layer \textit{convolutional} terakhir, teknik ini dapat juga digunakan pada semua layer \textit{convolutional}~\citep{gradcam} sehingga konsep \textit{explainability} dapat dilakukan secara menyeluruh.

\subsection{Penelitian Terkait}
\label{bab2:penelitianterdahulu}
Penelitian yang dilakukan oleh \citet{coral-cnn} menggunakan data citra yang diambil dari \textit{online databases} (RSMAS dan EILAT) yang terdiri atas tiga kelas, yaitu \textit{bleached coral, dead coral,} dan \textit{healthy coral}. Proses pemodelan CNN yang diusulkan oleh \citet{coral-cnn} dimulai dari \textit{setup environment} untuk model \textit{neural networks}. Kemudian, menentukan parameter yang digunakan oleh model (dimensi data citra, jumlah data \textit{training} dan validasi, jumlah epochs, dan \textit{batch size}. Terakhir, melakukan \textit{training} model dan melakukan prediksi citra untuk menentukan kelas terumbu karang.
\tabel~\ref{tab:hasil-prediksi} menyajikan performa model yang diusulkan oleh~\citet{coral-cnn}.
\begin{table}[!t]
    \centering
\begin{tabular}[t]{ccc}
\hline
\textbf{} & 
RSMAS & 
EILAT \\
\hline
\textbf{Jumlah Epoch} & 20 & 10 \\
\textbf{\textit{Training Accuracy}} & 75\% & 86,93\% \\
\textbf{\textit{Validation Accuracy}} & 68,75\% & 84,93\% \\
\textbf{Jumlah Test yang Benar} & 21/30 & 15/15 \\
\hline
\end{tabular}
    \caption{Hasil prediksi untuk setiap \emph{dataset}~\citep{coral-cnn}.}
    \label{tab:hasil-prediksi}
\end{table}


\section{Dataset}
\label{bab:3}

Pada bagian ini, akan dibahas tentang dataset yang digunakan pada pengerjaan ini. \subbab~\ref{bab3:gather} menjelaskan bagaimana cara dataset dikumpulkan. \subbab~\ref{bab3:analysis} mendiskusikan hasil analisis dataset. \subbab~\ref{bab3:distribution} menyampaikan distribusi data pada dataset, lalu memberikan \textit{insight} pada hasil distribusi tersebut.

\subsection{Pengumpulan Dataset}
\label{bab3:gather}
Dataset yang digunakan berasal dari Flickr menggunakan API Flickr, berupa citra terumbu karang di dalam laut yang diklasifikasikan menjadi dua kelas, yaitu terumbu karang yang dalam kondisi baik dan terumbu karang yang dalam kondisi rusak. Dataset ini terdiri dari 923 gambar terumbu karang, dimana setiap gambar telah diubah ukurannya agar memiliki dimensi maksimum 300 piksel untuk lebar atau tinggi, yang memastikan ukuran yang konsisten di seluruh dataset.

\subsection{Analisis Dataset}
\label{bab3:analysis}
\begin{figure*}
    \centering
    \begin{minipage}{0.32\textwidth}
        \centering
        \includegraphics[width=\textwidth]{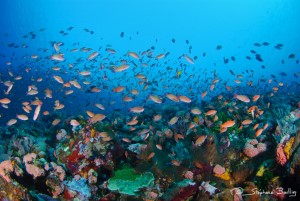}
        \subcaption{}
        \label{fig:tertutup_ikan}
    \end{minipage}
    \hfill
    \begin{minipage}{0.29\textwidth}
        \centering
        \includegraphics[width=\textwidth]{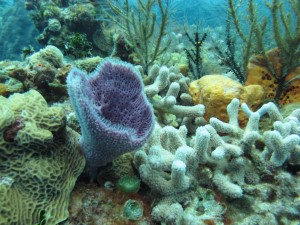}
        \subcaption{}
        \label{fig:bentuk_beda}
    \end{minipage}
    \hfill
    \begin{minipage}{0.33\textwidth}
        \centering
        \includegraphics[width=\textwidth]{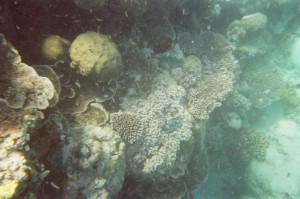}
        \subcaption{}
        \label{fig:blur_image}
    \end{minipage}
    \caption{Citra terumbu karang tertutup oleh ikan (a), berbeda bentuk (b), dan kurang jelas (c).}
    \label{fig:dataset}
\end{figure*}

Citra yang digunakan untuk melakukan \emph{training} memperhatikan berbagai kasus yang mungkin terjadi pada gambar \emph{dataset}, seperti terdapat ikan yang menutupi terumbu karang (\gambar~\ref{fig:tertutup_ikan}), bentuk terumbu karang berbeda-beda (\gambar~\ref{fig:bentuk_beda}), dan resolusi gambar kurang jelas (\gambar~\ref{fig:blur_image}).

\subsection{Distribusi Dataset}
\label{bab3:distribution}

\begin{table}[t!]
    \centering
\begin{tabular}[t]{cc}
\hline
     \textbf{Kelas} & 
     \textbf{Jumlah Sampel}\\ [1ex] 
     \hline
     
     \emph{Bleached Coral} & 485\\
     
     \emph{Healthy Coral} & 438\\ [1ex] 
     \hline
\end{tabular}
    \caption{
        Jumlah sampel masing-masing kelas.
        \label{tab:data-distribution}
    }
\end{table}

\emph{Dataset} memiliki 485 gambar terumbu karang yang dalam kondisi rusak, serta 438 gambar terumbu karang yang dalam kondisi baik (\tabel~\ref{tab:data-distribution}). Data ini sudah termasuk cukup \textit{balance}, sehingga dapat menghindari terjadinya bias pada model.

\section{Metodologi}
\label{bab:4}

Dari \emph{dataset} yang ada, dapat dilihat bahwa setiap citra memiliki banyak sekali informasi yang harus diekstrak, seperti ada citra ikan (\gambar~\ref{fig:tertutup_ikan}), \textit{healthy coral} dan \textit{bleached coral} yang saling berdekatan, serta bentuk terumbu karang yang berbeda-beda (\gambar~\ref{fig:bentuk_beda}). Hal ini dapat menyulitkan model untuk mengekstrak informasi yang ada pada citra karena citra yang rumit. Salah satu solusi untuk mengekstrak informasi dari citra yang rumit adalah dengan memperbanyak layer pada model \textit{neural network}. Namun, semakin banyak jumlah layer dapat menyebabkan \textit{loss information} dan \textit{error} yang tinggi pada saat \emph{training}. Oleh karena itu, digunakan model \textit{residual network} untuk mengatasi hal ini. Arsitektur ResNet memiliki jumlah \emph{layer} yang banyak untuk mengekstrak informasi yang rumit dan dapat mencegah \textit{loss information} pada saat \textit{training} dengan menambahkan \textit{shortcut connections} pada setiap \textit{building block} model.

\begin{figure}
    \centering
    \begin{tikzpicture}
        \node[draw, line width=0.5pt, inner sep=0pt] (image) at (0,0) {\includegraphics[width=0.47\textwidth]{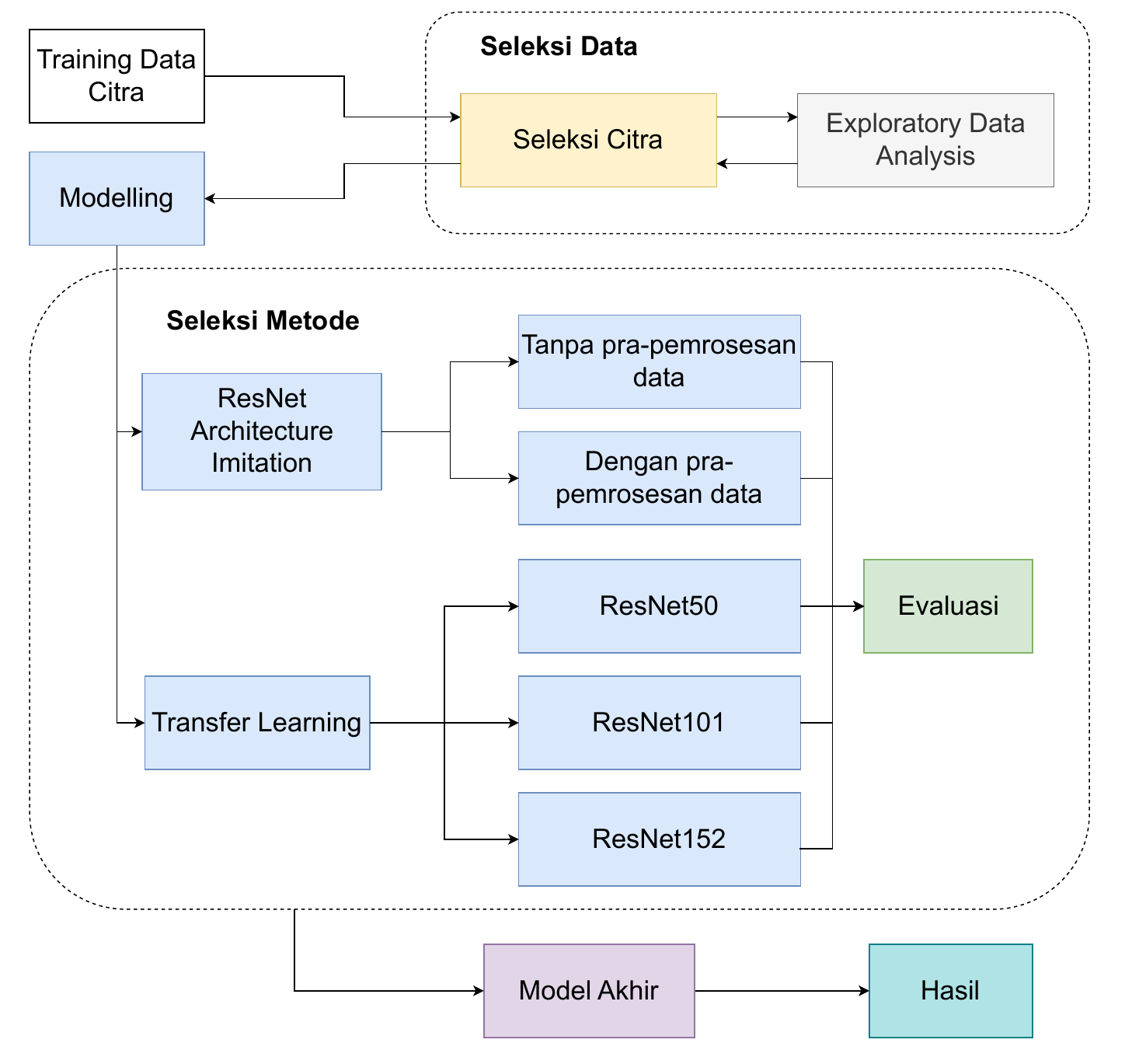}};
    \end{tikzpicture}
    \caption{
        Skema diagram alur eksperimen.
        \label{fig:metode-eksperimen}
    }
\end{figure}

Secara garis besar, langkah eksperimen yang akan tim peneliti lakukan sesuai dengan \gambar~\ref{fig:metode-eksperimen}.
Pada awalnya, data citra akan dianalisis terlebih dahulu untuk menyeleksi data yang relevan dan membuang data yang buruk seperti di bagian seleksi data pada \gambar~\ref{fig:metode-eksperimen}. Eksperimen klasifikasi citra \emph{coral} dilakukan dengan membandingkan hasil penggunaan model CNN yang mengimitasi arsitektur ResNet dan model hasil \textit{transfer learning} yang didapatkan dari \textit{package} tensorflow seperti di bagian seleksi metode pada \gambar~\ref{fig:metode-eksperimen}. Model-model \emph{transfer learning} yang digunakan adalah ResNet50, ResNet101, dan ResNet152. 

Pada model yang mengimitasi arsitektur ResNet dilakukan \textit{training} menggunakan dua data. Pertama, menggunakan data citra tanpa dilakukan pra-pemrosesan terlebih dahulu. Kedua, menggunakan data citra yang telah dipertajam (\textit{sharpened image}). Hal ini dilakukan untuk mendapatkan citra \emph{coral} yang bagian \textit{bleached}-nya lebih jelas (\gambar~\ref{fig:blur_image}). Kemudian, untuk model \textit{transfer learning} akan dilatih menggunakan data citra tanpa dilakukan pra-pemrosesan terlebih dahulu.

Setelah kelima proses \textit{training model} tersebut selesai. Setiap model hasil \textit{training} akan digunakan untuk memprediksi data validasi yang terdiri atas $231$ citra. Kemudian, hasil prediksi tersebut akan dievaluasi menggunakan \textit{metrics accuracy, precision} dan \textit{recall} untuk membandingkan performa kelima model.

\noindent{Kode hasil penelitian ini tersedia secara publik.}\footnote{\url{https://bit.ly/ThreeFoldsGemastik}}

\section{Eksperimen dan Analisis}
\label{bab:5}

Kami melakukan serangkaian pengujian dan analisis terhadap beberapa model ResNet untuk mengukur performa model-model tersebut dalam mendeteksi \emph{bleaching} pada citra terumbu karang. 

\begin{table}[H]
    \centering
{\small
\begin{tabular}{c c c c} 
     \hline
     \textbf{Model} & \textbf{\emph{Accuracy}} & \textbf{\emph{Precision}}\\ [1ex] 
     \hline
     ResNet101 & 0,72 & 0,81\\
     
     ResNet152 & 0,74 & 0,75\\ 
     
     ResNet50 & 0,72 & 0,73\\
     
     ResNet (\emph{Imitation}) & 0,78 & \textbf{0.82}\\ 
     
     ResNet (\emph{Imitation}) + \emph{preprocessing} & \textbf{0,79} & 0,76\\ [1ex] 
    
     \hline
\end{tabular}
}

     
     
     
     
    

    \caption{
        Tabel skor hasil pengujian dengan metrik \emph{accuracy} dan \emph{precision}.
        \label{tab:model-result}
    }
\end{table}

\begin{figure*}
    \centering
    \begin{minipage}{0.45\textwidth}
        \centering
        \includegraphics[width=\textwidth]{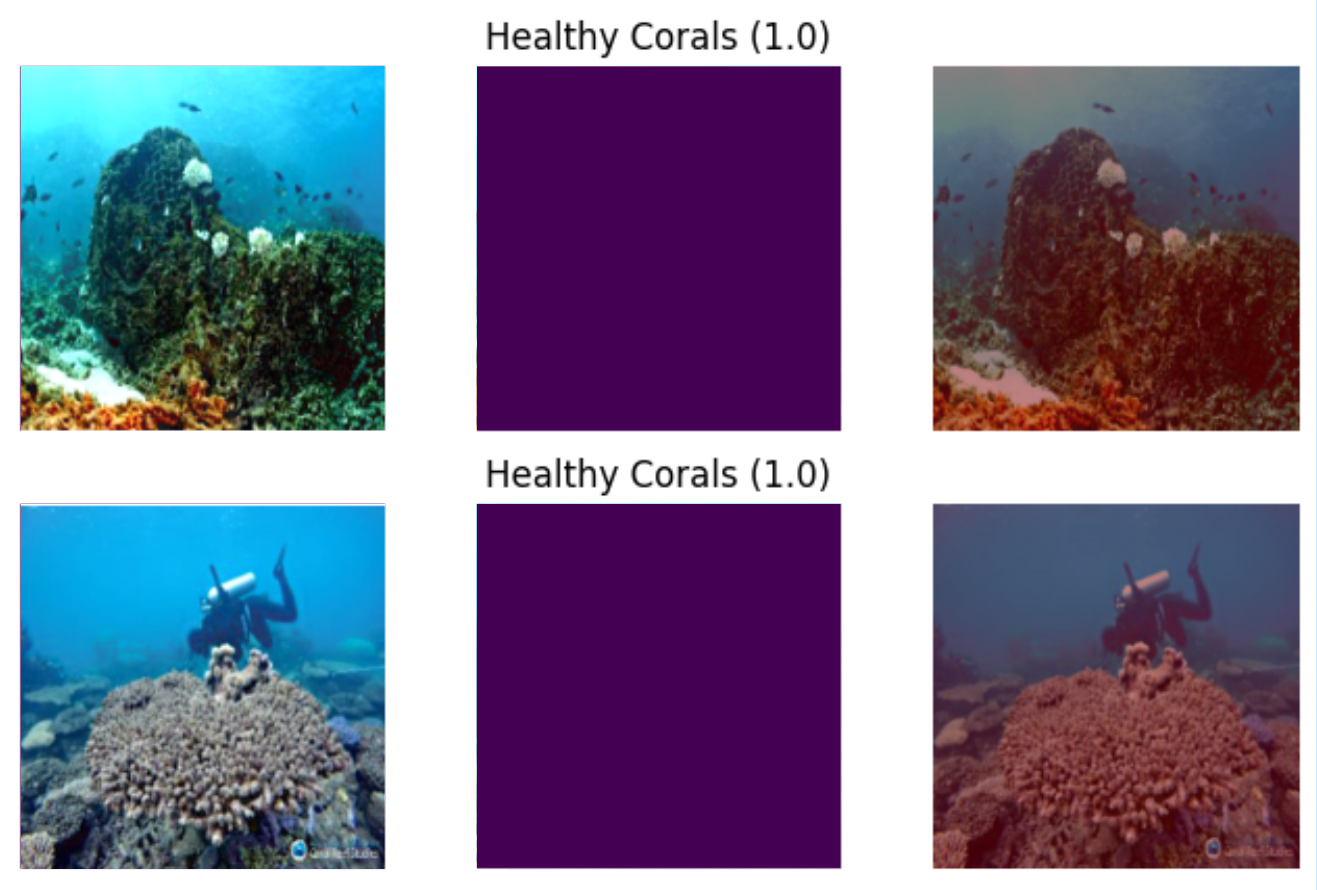}
        \subcaption{}
    \end{minipage}
    \hfill
    \begin{minipage}{0.45\textwidth}
        \centering
        \includegraphics[width=\textwidth]{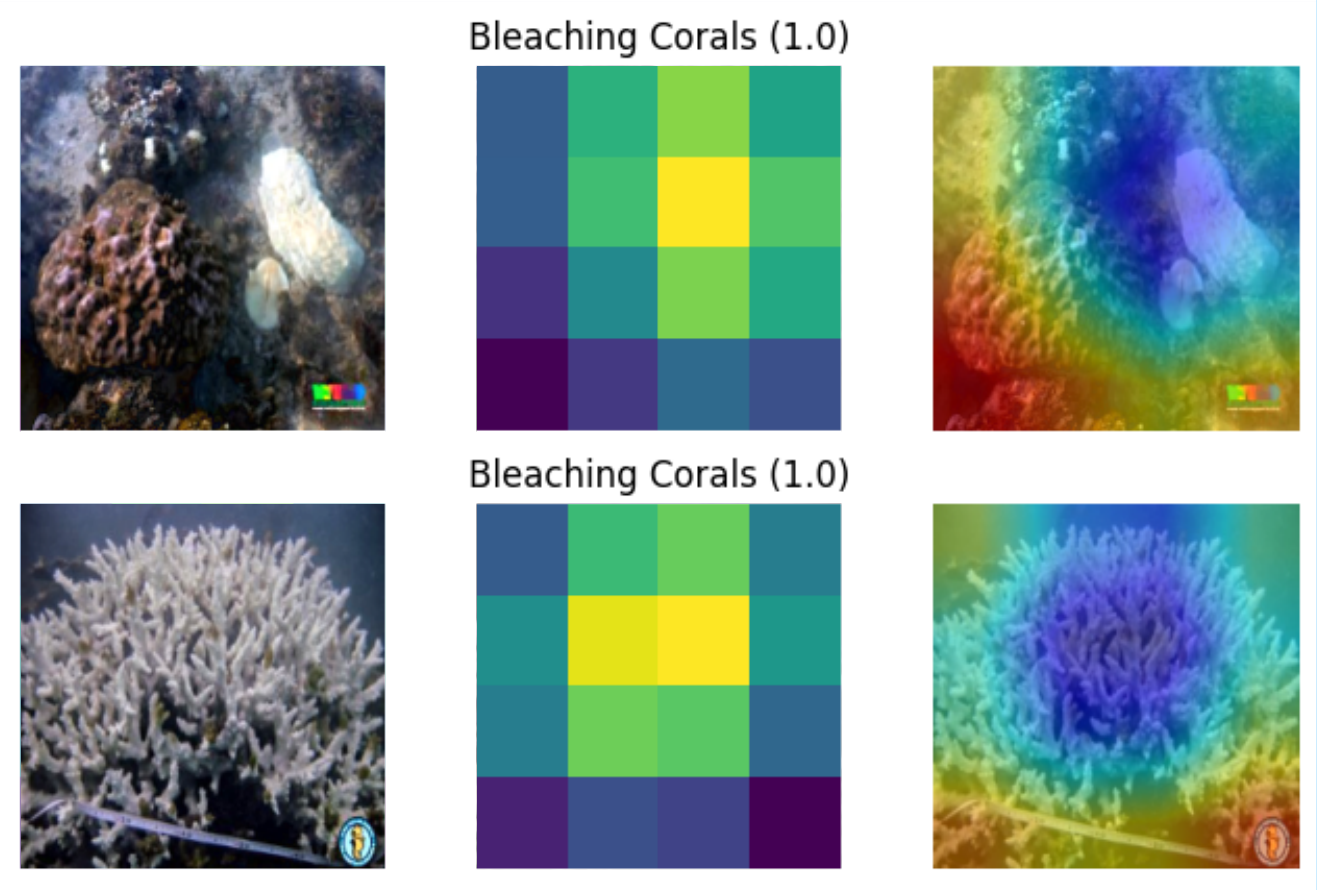}
        \subcaption{}
    \end{minipage}
    \caption{
        Grad-CAM hasil prediksi model untuk \textit{healthy coral} (a) dan \textit{bleached coral} (b) dengan 100\% \emph{confidence}.
        \label{fig:gradcam}
    }
\end{figure*}

Dari hasil pengujian, terlihat bahwa model ResNet \textit{imitation} menghasilkan skor paling tinggi untuk metrik \textit{accuracy}, serta untuk metrik \textit{precision} paling tinggi diperoleh dari model yang sama dengan tambahan \textit{preprocessing} pada data. Perlu diperhatikan bahwa ResNet \textit{imitation} adalah \textit{model CNN from scratch} yang menerapkan arsitektur \textit{residual network}, sedangkan model lainnya merupakan \textit{pretrained model} yang ditambahkan satu \textit{dense layer} dan satu \textit{output layer}.

Dalam pengerjaan ini, \textit{model from scratch} bekerja lebih baik daripada \textit{pretrained model} karena dataset yang relatif sedikit, sehingga menyebabkan \textit{pretrained model} menjadi \textit{overfit}. Jika diaplikasikan pada dataset lain yang cukup banyak, \textit{pretrained model} seharusnya bisa bekerja lebih baik.

Selain itu, \textit{preprocessing sharpen image} juga memiliki dampak pada bertambahnya skor metrik. Hal itu disebabkan karena kebanyakan dataset berlatar di dalam laut, mengakibatkan perubahan warna pada terumbu karang. Maka dari itu, \textit{sharpen image} menghilangkan nuansa kekuningan/kehijauan yang tidak diinginkan dan memberikan kejernihan pada air.

Untuk memvisualisasikan bagian-bagian gambar terumbu karang yang menjadi fokus model dalam membuat prediksi klasifikasi, digunakan algoritma Grad-CAM (Gradient-weighted Class Activation Mapping). Warna terang pada hasil Grad-CAM meunjukkan kontribusi tinggi terhadap prediksi kelas yang dituju, yaitu \textit{bleached}. Sebaliknya, warna gelap menunjukkan kontribusi sangat rendah atau tidak relevan terhadap prediksi kelas, yang artinya cenderung menuju ke kelas \textit{healthy}. Dapat dilihat pada \gambar~\ref{fig:gradcam}, hasil visualisasi Grad-CAM citra terumbu karang yang terdeteksi sebagai \emph{bleached} cenderung memiliki warna terang pada bagian citra yang mengindikasikan terjadinya \emph{bleaching}, sedangkan citra terumbu karang yang terdeteksi sehat cenderung memiliki warna gelap yang berarti tidak terdeteksi adanya \emph{bleaching}.



\section{Kesimpulan}


Kami sudah mengembangkan model yang dapat digunakan untuk mendeteksi apakah telah terjadi pemutihan terumbu karang melalui media citra bawah laut. Kami kemudian membangun data dari awal menggunakan API Flickr, sebelum akhirnya menggunakan konsep \emph{transfer learning} dan arsitektur ResNet untuk pengembangan model.

Implementasi \emph{deep learning} dengan memanfaatkan model ResNet telah membuktikan kemampuannya dalam melakukan pengklasifikasian terumbu karang yang rusak dengan tingkat keakuratan yang cukup baik. Namun, terdapat beberapa aspek yang dapat dieksplorasi lebih lanjut untuk meningkatkan performa sistem ini serta penerapannya dalam pengawasan terumbu karang secara efisien dan murah.

Salah satu aspek yang dapat ditingkatkan adalah memperluas \emph{dataset} yang digunakan dalam pelatihan model. Dengan menyertakan lebih banyak variasi citra terumbu karang yang rusak dan sehat, model dapat mempelajari pola-pola yang lebih kompleks dan memiliki kemampuan yang lebih baik dalam membedakan antara keduanya. Selain meningkatkan efektivitas model, peningkatan jumlah \emph{dataset} juga membantu mengatasi masalah \emph{overfitting} yang sering sekali terjadi.

Selain itu, \emph{preprocessing} pada citra dapat menjadi area peningkatan lainnya. Eksperimen sebelumnya telah menunjukkan bahwa penajaman warna pada gambar memiliki dampak terhadap performa model. Oleh karena itu, melakukan lebih banyak eksperimen \emph{preprocessing} pada citra, seperti penyesuaian warna yang lebih tepat, kontras, atau peningkatan resolusi, dapat membantu meningkatkan kemampuan sistem dalam mendeteksi kerusakan pada terumbu karang.

Dalam jangka panjang, sistem ini memiliki potensi untuk diimplementasikan dalam alat otomatis yang merekam kehidupan di bawah laut. Alat ini dapat secara terus-menerus memonitor kondisi terumbu karang dan mendeteksi secara otomatis jika ada kerusakan. Dengan adanya sistem ini, pengawasan terumbu karang menjadi lebih efisien dan murah dibandingkan dengan metode pengawasan manual. Hal ini dapat membantu konservasi terumbu karang dengan lebih baik dan mengurangi dampak negatif yang ditimbulkan oleh kerusakan terumbu karang, terutama dampak terhadap ekosistem.

Dalam rangka meningkatkan performa sistem dan menerapkan sistem ini, diperlukan upaya penelitian yang lebih lanjut. Selain itu, kerjasama antara ilmuwan, ahli kelautan, dan pemerintah juga penting untuk mendukung pengembangan dan penerapan teknologi ini secara luas. Dengan kerja sama yang baik, diharapkan pengawasan terumbu karang dapat dilakukan secara efektif dan berkelanjutan, sehingga terumbu karang dapat terjaga dan dilestarikan.

\pagebreak
\bibliographystyle{abbrvnat-mod}
\bibliography{Assets/refs}



\end{document}